\documentclass[]{SPAICE}
\usepackage{cuted} 
\usepackage{graphicx}
\usepackage{subcaption}
\usepackage{float}

\def\authorEmail{kaes@vision.rwth-aachen.de}
\def\AuthorShort{S. Käs et al.}

\author[]{Stephanie Käs  \thanks{Corresponding author. E-Mail: \authorEmail}}
\author[]{Marten Flaig}
\author[]{Lukas Woyke} 
\author[]{Osmanalp Ömeroğlu}
\author[]{Felix Krueckel}
\author[]{Onur Alp Çulha}
\author[]{Bastian Leibe}

\affil[]{Computer Vision Group, RWTH, Aachen, Germany}

\title{AstroPIE-53: Towards Understanding Failure Modes \\ of Human Pose Estimation in Space}

\begin{document}

\maketitle

\begin{abstract}
Reliable monitoring of astronauts is essential for safety, situational awareness, and autonomous assistance in space missions. While modern computer vision enables accurate human detection and pose estimation on Earth, its robustness in space environments remains largely unquantified due to limited data availability. We present AstroPIE-53, a pilot dataset of 53 annotated astronaut videos collected from publicly available mission footage. To respect copyright restrictions, we release video identifiers, reconstruction scripts, and dense annotations rather than raw videos. The dataset includes bounding boxes, challenge classes, and human-annotated pose failure tags. We benchmark state-of-the-art person detection and pose estimation models under realistic zero-shot deployment conditions and perform the first systematic failure analysis for astronaut monitoring. Our results reveal substantial performance degradation caused by occlusions, unusual body orientations, low contrast, and microgravity motion patterns.
\end{abstract}

\begin{figure}[t!]
    \centering
    \captionsetup[subfigure]{font=scriptsize} 
    \begin{tabular}{p{0.2\linewidth} p{0.2\linewidth} p{0.2\linewidth} p{0.2\linewidth}}
        \begin{subfigure}[b]{\linewidth}
            \centering
            \includegraphics[width=\linewidth, height=2cm, keepaspectratio=false]{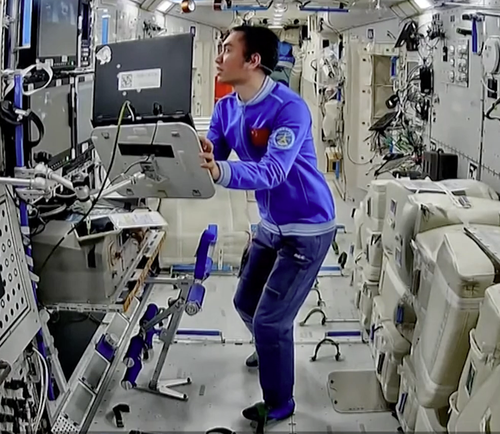}
            \caption{No suit (1)}
        \end{subfigure}
        &
        \begin{subfigure}[b]{\linewidth}
            \centering
            \includegraphics[width=\linewidth, height=2cm, keepaspectratio=false]{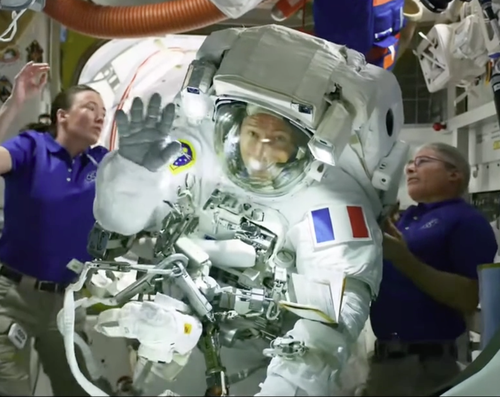}
            \caption{Suit (1)}
        \end{subfigure}
        &
        \begin{subfigure}[b]{\linewidth}
            \centering
            \includegraphics[width=\linewidth, height=2cm, keepaspectratio=false]{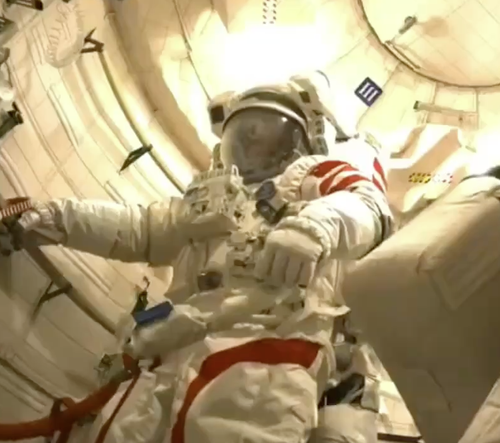}
            \caption{Indoor (3)}
        \end{subfigure}
        &
        \begin{subfigure}[b]{\linewidth}
            \centering
            \includegraphics[width=\linewidth, height=2cm, keepaspectratio=false]{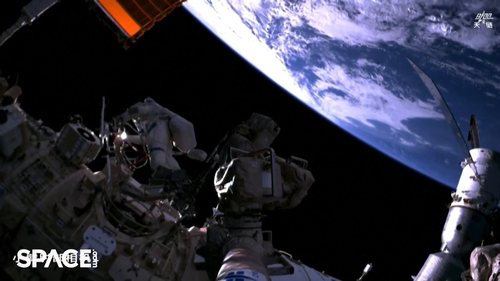}
            \caption{Outdoor (3)}
        \end{subfigure}
        \\[2ex]
        \begin{subfigure}[b]{\linewidth}
            \centering
            \includegraphics[width=\linewidth, height=2cm, keepaspectratio=false]{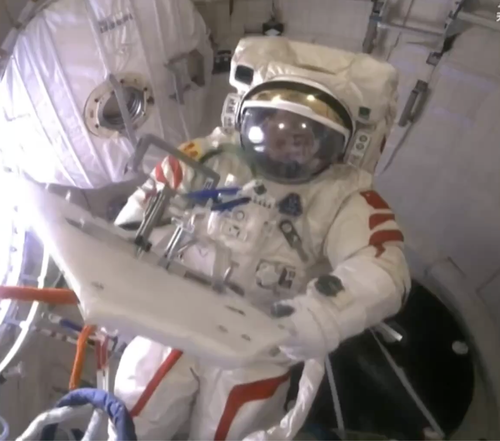}
            \caption{Front (2)}
        \end{subfigure}
        &
        \begin{subfigure}[b]{\linewidth}
            \centering
            \includegraphics[width=\linewidth, height=2cm, keepaspectratio=false]{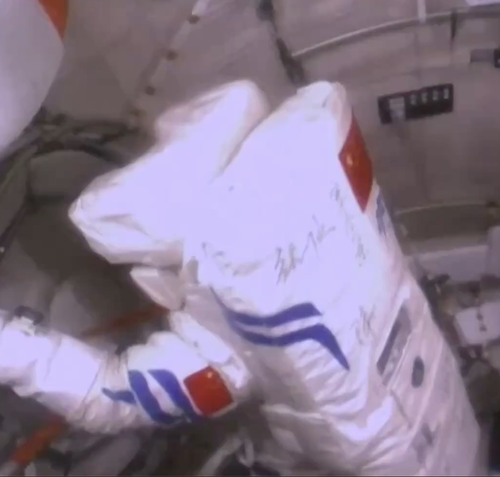}
            \caption{Back (2)}
        \end{subfigure}
        &
        \begin{subfigure}[b]{\linewidth}
            \centering
            \includegraphics[width=\linewidth, height=2cm, keepaspectratio=false]{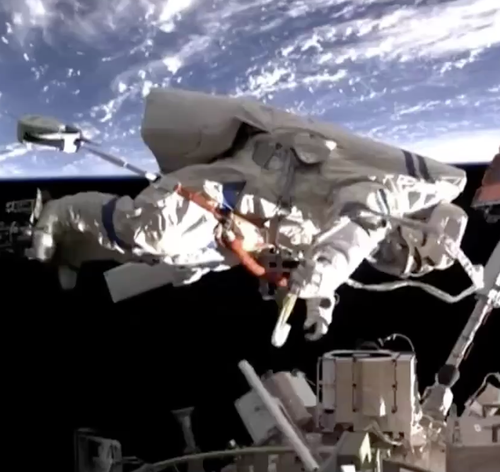}
            \caption{Side (2)}
        \end{subfigure}
        &
        \begin{subfigure}[b]{\linewidth}
            \centering
            \includegraphics[width=\linewidth, height=2cm, keepaspectratio=false]{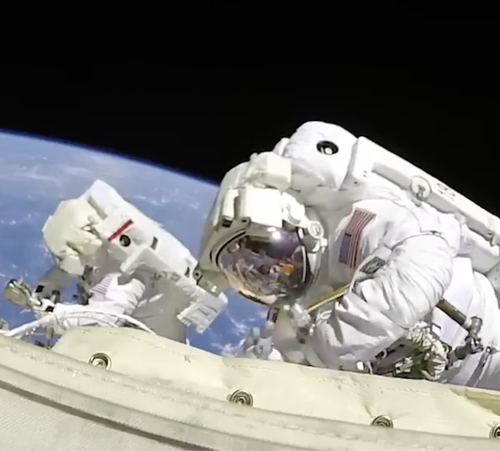}
            \caption{Multi (2)}
        \end{subfigure}
        \\[2ex]
        \begin{subfigure}[b]{\linewidth}
            \centering
            \includegraphics[width=\linewidth, height=2cm, keepaspectratio=false]{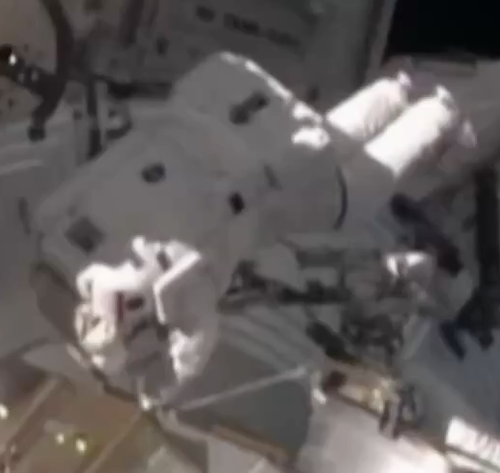}
            \caption{Low C. (4)}
        \end{subfigure}
        &
        \begin{subfigure}[b]{\linewidth}
            \centering
            \includegraphics[width=\linewidth, height=2cm, keepaspectratio=false]{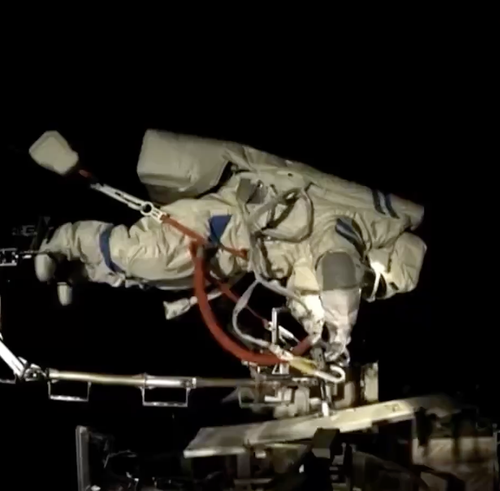}
            \caption{High C. (4)}
        \end{subfigure}
        &
        \begin{subfigure}[b]{\linewidth}
            \centering
            \includegraphics[width=\linewidth, height=2cm, keepaspectratio=false]{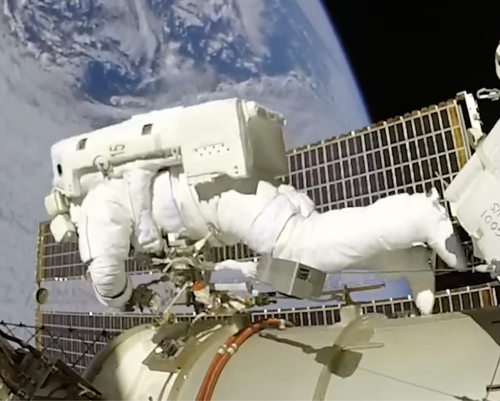}
            \caption{Close (5)}
        \end{subfigure}
        &
        \begin{subfigure}[b]{\linewidth}
            \centering
            \includegraphics[width=\linewidth, height=2cm, keepaspectratio=false]{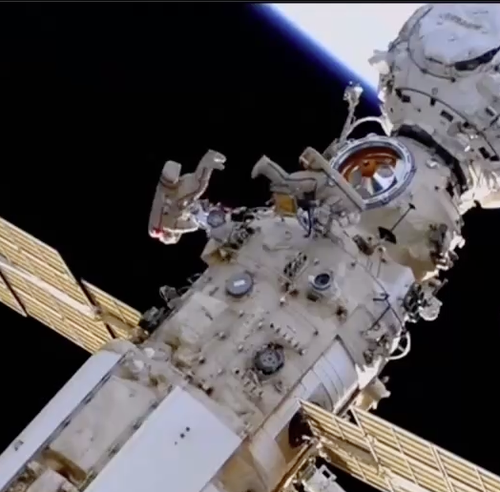}
            \caption{Far (5)}
        \end{subfigure}
        \\[2ex]
        \begin{subfigure}[b]{\linewidth}
            \centering
            \includegraphics[width=\linewidth, height=2cm, keepaspectratio=false]{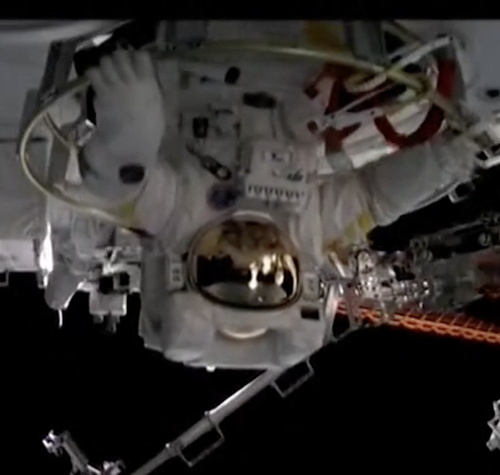}
            \caption{Occ. (6)}
        \end{subfigure}
        &
        \begin{subfigure}[b]{\linewidth}
            \centering
            \includegraphics[width=\linewidth, height=2cm, keepaspectratio=false]{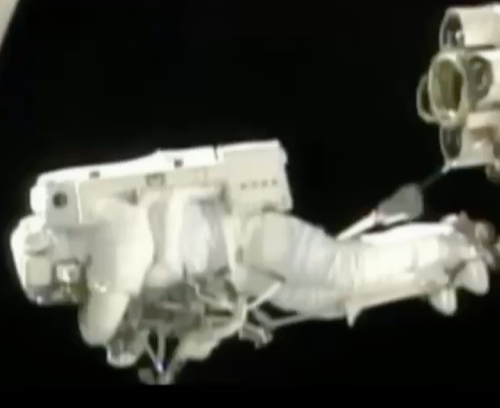}
            \caption{Non-Occ. (6)}
        \end{subfigure}
        &
        \multicolumn{2}{p{0.4\linewidth}}{
        \begin{minipage}[t]{\linewidth}
        \scriptsize
        \vspace{-10.2em}
        \textbf{Challenge Classes:}\\[0.3em]
        (1) Spacesuit Presence\\
        (3) Person-Camera Orientation\\
        (2) Recorded Environment\\
        (4) Person-Backgr. Contrast\\
        (5) Person-Camera Distance\\
        (6) Occlusion 
        \end{minipage}
        } \\
    \end{tabular}
    \caption{Example images from AstroPIE-53. We propose six different \textit{challenge classes} common in astronaut detection.}
    \label{fig:all_categories}
\end{figure}

\section{Introduction}
As human presence remains relevant for space exploration, ensuring astronaut safety is crucial for future missions. Autonomous monitoring of astronaut activity enables  health assessment, hazard detection, and human–robot collaboration. Several studies have investigated astronaut motion using marker-based optical motion capture \cite{Abercromby2006ReachEnvelope, Kobrick2018SpacesuitROM, Kobrick2019FinalFrontierROM, Doule2021IVASpacesuit} wearable inertial sensors \cite{Kobrick2012IMUSpacesuit, Fineman2018SpacesuitFit}, or combinations of both \cite{Kobrick2012IMUSpacesuit}. These approaches require additional body-worn devices which increase system complexity. Human pose estimation relying solely on onboard cameras without any additional sensors would be beneficial. Modern computer vision methods provide markerless, monocular approaches for accurate 3D human pose estimation in terrestrial environments \cite{HPE:Metrabs:Sarandi, ours:nlf}. However, their robustness under real space conditions remains largely unexplored. Space environments introduce severe domain shifts compared to standard vision datasets: unique visual characteristics including microgravity body orientations, bulky spacesuits, occlusions, and extreme image contrast variability. Publicly available astronaut imagery is scarce due to legal and operational constraints.

\begin{figure}[t!]
    \centering
    \includegraphics[width=0.90\linewidth]{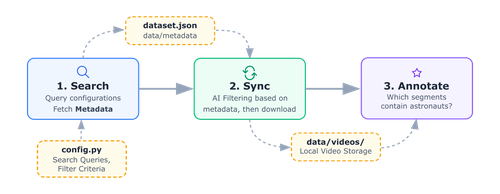}
    \caption{Three-stage data pipeline for constructing AstroPIE-53: (1) metadata-driven video search, (2) filtering and synchronization of relevant recordings, and (3) frame extraction and annotation of astronaut activities.}
    \label{fig:data_pipeline}
\end{figure}

\begin{figure*}[t!]
  \begin{minipage}{\textwidth}
    \centering

    \begin{minipage}[t]{0.30\linewidth}
    \vspace{0.55cm}
      \centering
      \captionsetup{type=figure}

      \begin{subfigure}[t]{\linewidth}
        \centering
        \includegraphics[
          width=\linewidth,
          height=0.20\textheight,
          keepaspectratio, 
          angle=90
        ]{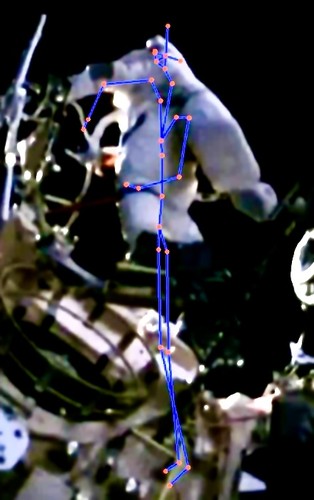}
        \caption{MeTRAbs \cite{HPE:Metrabs:Sarandi}} %
        \label{fig:error-metrabs}
      \end{subfigure}

      \vspace{0.5em}

      \begin{subfigure}[t]{\linewidth}
        \centering
        \includegraphics[
          width=\linewidth,
          height=0.14\textheight,
          keepaspectratio
        ]{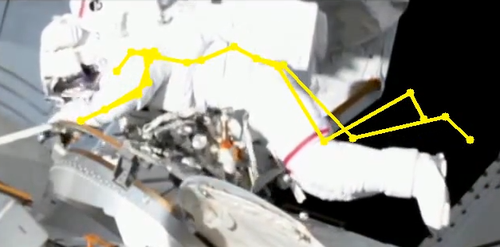}
        \caption{NLF \cite{data:nasa-challenge:cc20}} %
        \label{fig:error-nlf}
      \end{subfigure}

      \caption{Example of human pose estimation failure cases.  MeTRAbs \cite{HPE:Metrabs:Sarandi} output is completely misaligned; (b) NLF \cite{ours:nlf} output shows misaligned legs.}
      \label{fig:pipeline-left}
    \end{minipage}
    \hfill
    \begin{minipage}[t]{0.65\linewidth}
      \centering
      \captionsetup{type=figure}

      \includegraphics[
        width=\linewidth,
        height=0.40\textheight,
        keepaspectratio
      ]{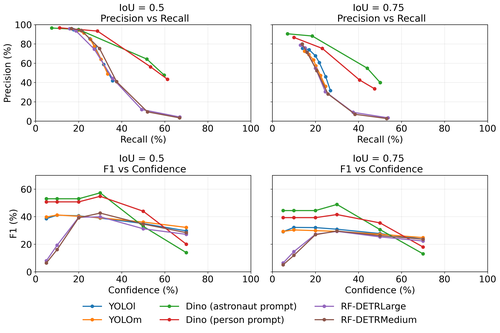}

      \caption{Precision–recall curves (top) and F1 score versus confidence threshold (bottom) for selected object detection models (YOLO26~\cite{det:yolo26:jocher}, RF-DETR~\cite{det:rf-detr:robinson}, Grounding DINO~\cite{det:grounding-dino:liu}) evaluated at IoU thresholds of 0.5 and 0.75 evaluated on AstroPIE-53. Results show a sweet spot around $\tau_{\text{conf}}=0.3$ and $\tau_{\text{IoU}}=0.5$.}
      \label{fig:pr}
    
    \end{minipage}

  \end{minipage}
\end{figure*}

No prior work systematically provides a quantitative analysis of how modern detection and pose estimation pipelines perform on authentic astronaut mission footage. We address this gap by (1) presenting AstroPIE-53, a pilot dataset of real astronaut videos with bounding box annotations, structured challenge classes, and pose failure tags.
(2) Using AstroPIE-53, we benchmark modern person detection and pose estimation models on authentic mission footage. (3) We then propose a taxonomy of space-specific visual challenges and quantify their impact on detection and pose estimation failures.

\section{Related Work}

\textbf{Human Pose Estimation.}
Vision-based human pose estimation (HPE) reconstructs human skeletons or meshes from images. Modern pipelines typically follow a two-stage approach: (1) person localization using object detectors (such as YOLO~\cite{det:yolo:redmon}, DETR~\cite{det:rf-detr:robinson} ), foundation models (like DINO~\cite{det:grounding-dino:liu}, Gemini~\cite{other:geminifamily:google}) which create bounding boxes around the person or body segmentation masks~\cite{meta_sam_3d}. (2) Using these body selections, pose estimation methods (such as MeTRAbs~\cite{HPE:Metrabs:Sarandi}, NLF~\cite{ours:nlf}, and ViTPos~\cite{hpe:vitpose:xu}) reconstruct human models. Other methods - like OpenPose~\cite{hpe:openpose:cao} - directly estimate potential body joints and assign them to persons afterwards. Skeletons/meshes can then be used for tasks such as gesture/activity recognition ~\cite{ours:nugget}.

\textbf{Astronaut Monitoring} Most existing approaches  (Table~\ref{tab:related_work}) use a single camera (\textit{monocular}). AstroPose ~\cite{hpe:astropose:liu} formulates pose estimation without the use of AI as a geometric optimization problem. Later works adopt CNN-based pipelines: SpacesuitPose~\cite{data:spacesuitpose:ouyang} enhances HRNet~\cite{hpe:hrnet:sun} with occlusion-aware augmentation and structural losses, $\mu$Pose ~\cite{data:mupose:cardoso} demonstrates improvements through synthetic microgravity fine-tuning, and ROpenPose~\cite{hpe:ropenpose:wu} proposes a lightweight OpenPose~\cite{hpe:openpose:cao} variant optimized for real-time posture tracking. H3-AstroPoseNet~\cite{hpe:h3-astroposenet:xun} uses multi-camera views, but lacks evaluation on space-like data. 

\textbf{Relevant Datasets.} Though many terrestrial HPE datasets exist (e.g. COCO~\cite{data:coco:lin}, EgoExo4D~\cite{data:egoexo4d:grauman}), existing astronaut HPE datasets are small and limited in diversity and availability (\cref{tab:related_work}). SpacesuitPose provides 400 annotated frames from EVA and training footage, while NASA’s Spacesuit Detection Challenge~\cite{data:nasa-challenge:cc20} released a small dataset that is no longer publicly accessible. AstroPose combines marker-based recordings with unannotated mission footage. $\mu$Pose introduces synthetic microgravity avatars in a single ISS-like environment, and OpenPose contributes an indoor capsule dataset focused mainly on posture classification. Spacewalk-18 ~\cite{data:spacewalk-18:tang}  introduces a long-form video understanding benchmark for space operation scenarios, but does not focus on person detection, pose estimation nor does it provide a failure analysis. In contrast, our dataset AstroPIE-53 includes more than 20k annotated frames with 2D bounding boxes and pose failure cases.

\textbf{Failure Sources.}
Prior work mentions challenges such as occlusions, bulky suits, lighting variability, and microgravity motion patterns. However, these factors are rarely quantified, and no study quantitatively analyzes failure sources across the full detection–pose pipeline. We address this gap by presenting a quantitative failure analysis of pose detection and pose estimation models on real astronaut recordings using our AstroPIE-53 dataset.

\begin{table*}
\centering
\caption{Overview of monocular methods and datasets related to astronaut monitoring. Our AstroPIE-53 benchmark includes a detailed failure case analysis regarding occlusion, person-camera distance, spacesuit existence, etc., whilst existing publications do not provide systematic quantification across detection and pose estimation. Some datasets are recorded in labs, others provide real data (inside/outside spacecraft (SC))}
\label{tab:datasets_failures}
\resizebox{\textwidth}{!}{
\begin{tabular}{lcccccc|cccccc}
\hline
 &  &  &  &  &  &  & \multicolumn{6}{c}{\textbf{Analyzed Failure Cases}} \\
\cline{8-13}
\textbf{Dataset / Work} & \textbf{Real} & \textbf{\#Frames} & \textbf{2D GT} & \textbf{3D GT} & \textbf{Env.} & \textbf{Public} 
& \textbf{Occl.} & \textbf{Dist.} & \textbf{Suit} & \textbf{Contr.} & \textbf{Microg.} & \textbf{Quant.} \\
\hline
SpacesuitPose~\cite{data:spacesuitpose:ouyang}& \checkmark & 400 & 19 joints & -- & Lab  & Yes 
& \checkmark & -- & \checkmark & \checkmark & -- & No \\

NASA Challenge~\cite{data:nasa-challenge:cc20} & \checkmark & $\sim$200 + video & \checkmark & \checkmark & Mixed & No 
& -- & -- &  \checkmark & -- & -- & No \\

AstroPose~\cite{hpe:astropose:liu} & Astron. Figure & -- & -- & -- & Lab & No 
& -- & -- &  \checkmark & -- & -- & No \\

$\mu$Pose~\cite{data:mupose:cardoso} & Synthetic & 312 (aug.) & \checkmark & -- & Inside SC & Yes 
& -- & -- & -- & -- & \checkmark & No \\

ROpenPose~\cite{hpe:ropenpose:wu} & \checkmark & $\sim$9k & Unclear & -- & Inside SC & No 
& \checkmark & -- & -- & -- & \checkmark & No \\

Spacewalk-18~\cite{data:spacewalk-18:tang} & \checkmark & $\sim$10M & -- & -- & Outside SC & Yes
& -- & -- & -- & -- & -- & No \\

\textbf{AstroPIE-53 (ours)} & \checkmark & $\sim$20.5k & -- & -- & Mixed & Upon acc. 
& \checkmark & \checkmark & \checkmark & \checkmark & \checkmark & Yes \\
\hline
\end{tabular}
}
\label{tab:related_work}
\end{table*}

\section{AstroPIE-53 Dataset}

We introduce \textbf{AstroPIE-53}, a pilot dataset designed to analyze failure modes in vision-based astronaut monitoring from real mission footage.

\textbf{Data Collection:}
We collected 53 videos from publicly available astronaut recordings published by space agencies. The footage spans intra-vehicular and extra-vehicular activities, including maintenance tasks, tours, and mission operations. To construct the dataset, we developed a semi-automated three-stage pipeline consisting of \textit{Search}, \textit{Synchronization}, and \textit{Annotation} (see \cref{fig:data_pipeline}). In the \textit{Search} stage, candidate videos are identified using keyword-based queries and stored as structured metadata (video IDs, titles, timestamps, and URLs). The \textit{Sync} stage filters and downloads relevant videos while ensuring consistency between the metadata repository and the local dataset. Finally, the \textit{Annotate} stage processes the synchronized videos to extract frames and generate annotations. 
To respect copyright restrictions, raw videos are not redistributed. Instead, we release reconstruction scripts, video identifiers, and annotations, enabling researchers to deterministically reconstruct the dataset from publicly available sources.

\textbf{Annotations:}
We manually annotated three types of labels using a custom GUI: 
(1) \textit{Bounding Boxes:}
Manual astronaut detection annotations (20.5k frames).(2) \textit{Failure Tags:}
Human-annotated error categories describing failures of outputs. Failure tags are assigned per astronaut track and may be multi-label.  (3) \textit{Challenge Classes:} Environmental tags including occlusion, low illumination, suit presence, person-camera orientation, and camera distance. We structure AstroPIE-53 accordingly (see \cref{tab:dataset_stats}).

\begin{table*}[t]
\centering
\footnotesize
\caption{Statistics of our AstroPIE-53 dataset. It is divided into 14 different challenge label subsets across six challenge classes.} %
\label{tab:dataset_stats}
\resizebox{\textwidth}{!}{
\begin{tabular}{lcccccccccccccc}
\toprule
 & \multicolumn{2}{c}{Spacesuit} & \multicolumn{4}{c}{Orientation} & \multicolumn{2}{c}{Env.} & \multicolumn{2}{c}{Contrast} & \multicolumn{2}{c}{Distance} & \multicolumn{2}{c}{Occlusion} \\
\cmidrule(lr){2-3} \cmidrule(lr){4-7} \cmidrule(lr){8-9} \cmidrule(lr){10-11} \cmidrule(lr){12-13} \cmidrule(lr){14-15}
 & Suit & No-Suit & Front & Back & Side & Multi & Ind. & Out. & High & Low & Close & Far & Occ. & Non-Occ. \\
\midrule
Videos & 43 & 10 & 7 & 3 & 25 & 18 & 15 & 38 & 20 & 18 & 46 & 7 & 9 & 44 \\
Frames [\%] & 86.0 & 14.0 & 8.4 & 7.0 & 34.6 & 50.0 & 21.0 & 79.0 & 39.3 & 60.7 & 68.8 & 31.2 & 17.1 & 82.9 \\
\bottomrule
\end{tabular}
}
\end{table*}

\section{Methodology}

We analyze failure modes of vision-based astronaut monitoring using a representative detection–pose pipeline, providing a broad coverage of state-of-the-art detection and pose estimation strategies.

\textbf{Person Detection:} We evaluate three complementary object detection paradigms: YOLO26~\cite{det:yolo26:jocher} represents high-speed convolutional detectors commonly used in real-time robotics applications. RF-DETR~\cite{det:rf-detr:robinson} is a transformer-based detector designed for robust localization under scale variation and clutter. Grounding DINO~\cite{det:grounding-dino:liu} is a foundation detection model that leverages vision–language alignment and enables open-vocabulary detection via text prompts. These models were selected to cover the dominant detection paradigms currently deployed in real-world perception systems, ranging from conventional CNN detectors to modern transformer and foundation-model approaches.

\textbf{Pose Estimation:} For pose estimation, we evaluate three representative top-down methods: CNN-based MeTRAbs~\cite{HPE:Metrabs:Sarandi} performs direct regression of 3D human skeletons and is known for robustness to viewpoint variation and compatibility with fisheye cameras~\cite{Ours:fishnchips}. NLF~\cite{ours:nlf} reconstructs parametric human meshes using neural body models, enabling detailed structural predictions. ViTPose~\cite{hpe:vitpose:xu} is transformer–based using global attention~\cite{Vaswani2017Attention} to predict body keypoints. 

\section{Experiments}

We evaluate representative state-of-the-art models on AstroPIE-53 and its challenge class subsets (without retraining) to simulate realistic deployment conditions. 

\subsection{Astronaut Detection per Challenge Class}

We benchmark Grounding DINO~\cite{det:grounding-dino:liu}, RF-DETR~\cite{det:rf-detr:robinson}, and YOLO26~\cite{det:yolo26:jocher} on astronaut detection using AstroPIE-53 challenge subsets.  RF-DETR~\cite{det:rf-detr:robinson} and YOLO26~\cite{det:yolo26:jocher} rely on fixed label sets. Only detections classified as \textit{person} are considered, as there is no \textit{astronaut} class.  Grounding DINO uses text prompts and is evaluated using both \textit{astronaut} and \textit{person} with a text threshold of $\tau_{\text{text}}=0.25$. As an evaluation metric, we use the AP@50 metric, measuring the \textit{Average Precision at an  Intersection-over-Union (IoU) threshold $\tau_{\text{IoU}}=0.5$} and report the results in \cref{tab:eval_pd}. We selected the confidence thresholds ($\tau_{\text{conf}}=0.3$) based on our ablation studies (\cref{sec:ablation}).

\textbf{Results \& Interpretation:} As seen in \cref{tab:eval_pd}, detection performance is substantially higher without spacesuits among all three detectors. Frontal views yield the best  (DINO: $\sim$95\%, Others: $\sim$86\%), whereas back views significantly reduce performance (All: $\leq$25\%, YOLO: $\sim$2\%), with side and multi-view cases in between. This might be caused by large backpacks carried by astronauts. Environmental conditions follow a similar pattern to viewing angles since outdoor recordings exclusively contain astronauts wearing spacesuits. Image contrast strongly affects performance: high contrast improves AP@50, while low contrast degrades it  (All: $\leq$25\%). Large camera distances lead to the strongest performance drop  (DINO: $\leq$23\%, Others: $\leq$1\%) and represent the most challenging scenario. Unexpectedly, occluded frames yield higher AP, likely reflecting viewpoint, distance and contrast rather than occlusion itself. The comparison clearly shows that DINO significantly outperforms YOLO26 and RF-DETR on AstroPIE-53, which basically fail to detect anything in outdoor scenarios, at low contrast and far distances.

\subsection{Ablation Study on Detector Thresholds}
\label{sec:ablation}
We compare different $\tau_{\text{conf}}$ and $\tau_{\text{IoU}}$ for Grounding DINO~\cite{det:grounding-dino:liu}, RF-DETR~\cite{det:rf-detr:robinson}, and YOLO26~\cite{det:rf-detr:robinson} on astronaut detection using AstroPIE-53.

\textbf{Results:} \cref{fig:pr} shows precision–recall and F1–confidence curves for IoU thresholds of 0.5 and 0.75. Grounding DINO with the astronaut prompt achieves the highest F1 score at both IoU levels, peaking around $\tau_{\text{conf}}=0.3$. The person prompt follows a similar trend but with slightly lower performance. YOLO variants exhibit more stable behavior across confidence thresholds, particularly at higher confidence levels, where their F1 scores remain comparatively consistent. In contrast, RF-DETR variants show low precision at $\tau_{\text{conf}}\leq0.2$ and overall lower F1 scores. Performance decreases when increasing the $\tau_{\text{IoU}}$ from 0.5 to 0.75, but the relative ranking of the models remains largely consistent.

\begin{table}[t]
\centering
\footnotesize
\caption{AP@50 (\%) of tested person detectors ( $\tau_{\text{conf}}=0.3$, $\tau_{\text{IoU}}=0.5$) on AstroPIE-53 evaluated per challenge category. Detection performance drops sharply for suit presence/outdoors, back views, low contrast and far distances. Overlapping, imbalanced categories yield associations, not isolated effects.}
\resizebox{\columnwidth}{!}{
\begin{tabular}{lcccc}
\toprule
 & DINO~\cite{det:grounding-dino:liu}& DINO~\cite{det:grounding-dino:liu} & RF-DETR~\cite{det:rf-detr:robinson} & YOLO26~\cite{det:yolo26:jocher} \\
\textbf{Categories [\%]} & (Astronaut) & (Person) & &  \\
\midrule
Suit        & 46.03 & 46.88 & 16.78 & 14.85 \\
No-Suit     & 86.90 & 86.87 & 85.76 & 84.40 \\
\midrule
Front       & 94.91 & 94.57 & 86.15 & 86.22 \\
Back        & 13.62 & 24.64 & 6.52  & 2.18  \\
Side        & 76.24 & 73.93 & 30.17 & 23.53 \\
Multi       & 31.91 & 33.45 & 14.81 & 16.30 \\
\midrule
Indoor      & 81.00 & 85.92 & 80.80 & 80.99 \\
Outdoor     & 43.66 & 43.37 & 11.70 & 9.34  \\
\midrule
High Contrast & 73.96 & 68.94 & 22.66 & 18.40 \\
Low Contrast  & 24.05 & 26.82 & 4.61  & 3.48  \\
\midrule
Close       & 66.67 & 65.96 & 37.46 & 34.45 \\
Far         & 18.42 & 22.21 & 0.23  & 0.83  \\
\midrule
Occluded    & 52.15 & 52.76 & 27.87 & 25.70 \\
Non-Occluded & 46.66 & 47.99 & 14.00 & 13.72 \\
\bottomrule
\end{tabular}
}
\label{tab:eval_pd}
\end{table}

\subsection{Pose Failure Analysis}

Since ground-truth pose annotations are unavailable for real mission footage, we perform a structured human-annotated failure analysis rather than reporting classical pose accuracy metrics.
For each astronaut track, predicted pose sequences from MeTRAbs~\cite{HPE:Metrabs:Sarandi}, NLF~\cite{ours:nlf}, and ViTPose~\cite{hpe:vitpose:xu} are labeled with predefined failure categories: misaligned body parts, incorrect scale, wrong global orientation, fully misaligned poses, temporal jitter, and shadowing. We use AstroPIE-53 ground-truth bounding boxes to isolate pose estimation errors from detection errors.

\begin{table}[t]
\centering
\footnotesize
\caption{Human-annotated failure analysis of three pose estimators on AstroPIE-53. Each cell reports the percentage of 19912 annotated bounding-box keyframes exhibiting the respective failure type, with respect to 20489 total video frames. As multiple misalignments can occur on the same bounding box frame the total sum of percentages will not necessary add up to 100\%. Correct Estimations denotes keyframes with no failure tag.}
\setlength{\tabcolsep}{8pt}
\renewcommand{\arraystretch}{0.85}
\begin{tabular}{@{}lccc@{}}
\toprule
Failure Type [\%] & MeTRAbs & NLF & ViTPose \\
\midrule
Misaligned head & 3.9 & --- & --- \\
Misaligned torso & 0.9 & 0.1 & --- \\
Misaligned arms & 1.3 & 4.6 & 0.1 \\
Misaligned legs & 8.5 & 7.9 & 1.5 \\
Scale & --- & 3.4 & --- \\
Global orientation & 1.8 & 3.8 & 0.3 \\
All misaligned & 43.3 & 30.9 & 57.2 \\
Jitter & 4.9 & 1.9 & 9.8 \\
Shadowing & 1.0 & --- & 1.5 \\
Correct Estimations & 55.4 & 69.8 & 48.3 \\
\bottomrule
\end{tabular}
\label{tab:failure-cases}
\end{table}

\textbf{Results.} Table~\ref{tab:failure-cases} reports failure prevalence as a percentage of annotated keyframes.
Fully misaligned poses are the dominant failure, affecting 57.2\% (ViTPose), 43.3\% (MeTRAbs), and 30.9\% (NLF) of keyframes, potentially due to microgravity-related unusual body orientations.
VitPose exhibits the highest jitter rate (9.8\%), while NLF is affected by scale (3.4\%) and global orientation errors (3.8\%). Leg misalignment is the most common partial-body failure, peaking at 8.5\% for MeTRAbs. NLF provides the highest
proportion of correct estimates, with 69.8\% of frames being correctly annotated, while ViTPose provides the lowest accuracy at 48.3\%. No pose estimator dominates across all failure modes.

\section{Conclusion}

We introduce AstroPIE-53, the first dataset enabling systematic evaluation of vision-based astronaut monitoring. Our experiments demonstrate significant performance degradation of modern detection and pose estimation pipelines on authentic mission footage. By providing a reproducible dataset curation pipeline, systematic taxonomy of challenges for astronaut detection in space, and quantitative failure analysis, this work aims to draw attention to open research questions in astronaut monitoring. 

\printbibliography

@Misc{data:nasa-challenge:cc20,
  author       = "{{National Aeronautics and Space Administration}}",
  title        = "{CC20 NASA} Spacesuit Detection Challenge",
  howpublished = "NASA TechPort, Project 117518",
  year         = 2020,
  url          = "https://techport.nasa.gov/projects/117518",
  urldate      = "2026-08-11"
}

@inproceedings{Vaswani2017Attention,
  author    = {Vaswani, Ashish and Shazeer, Noam and Parmar, Niki and Uszkoreit, Jakob and Jones, Llion and Gomez, Aidan N. and Kaiser, Lukasz and Polosukhin, Illia},
  title     = {Attention Is All You Need},
  booktitle = {Advances in Neural Information Processing Systems},
  volume    = {30},
  year      = {2017},
  pages     = {5998--6008}
}

@inproceedings{meta_sam_3d,
  author    = {Yang, Xitong and Kukreja, Devansh and Pinkus, Don and Sagar, Anushka and Fan, Taosha and Park, Jinhyung and Shin, Soyong and Cao, Jinkun and Liu, Jiawei and Ugrinovic, Nicolas and Feiszli, Matt and Malik, Jitendra and Doll{\'a}r, Piotr and Kitani, Kris},
  title     = {{SAM} 3D Body: Robust Full-Body Human Mesh Recovery},
  booktitle = {2026 IEEE/CVF Conference on Computer Vision and Pattern Recognition (CVPR)},
  year      = {2026},
  pages     = {7209--7219},
  url       = {https://openaccess.thecvf.com/content/CVPR2026/papers/Yang_SAM_3D_Body_Robust_Full-Body_Human_Mesh_Recovery_CVPR_2026_paper.pdf}
}

@article{HPE:Metrabs:Sarandi,
  author  = {Istv{\'{a}}n S{\'{a}}r{\'{a}}ndi and
                  Timm Linder and
                  Kai Oliver Arras and
                  Bastian Leibe},
  title   = {{MeTRAbs}: Metric-Scale Truncation-Robust Heatmaps for Absolute 3{D} Human Pose Estimation},
  journal = {IEEE Transactions on Biometrics, Behavior, and Identity Science},
  year    = {2021},
  volume  = {3},
  number  = {1},
  pages   = {16--30},
  doi     = {10.1109/TBIOM.2020.3037257}
}

@inproceedings{Ours:fishnchips,
  author    = {K{\"a}s, Stephanie and Peter, Sven and Thillmann, Henrik and Burenko, Anton and Adrian, David Benjamin and Mack, Dennis and Linder, Timm and Leibe, Bastian},
  title     = {Systematic Comparison of Projection Methods for Monocular 3{D} Human Pose Estimation on Fisheye Images},
  booktitle = {2025 IEEE International Conference on Robotics and Automation (ICRA)},
  year      = {2025},
  pages     = {5914--5921},
  doi       = {10.1109/ICRA55743.2025.11128613}
}

@misc{other:geminifamily:google,
  author        = {{Gemini Team, Google}},
  title         = {{Gemini}: A Family of Highly Capable Multimodal Models},
  year          = {2023},
  eprint        = {2312.11805},
  archiveprefix = {arXiv},
  primaryclass  = {cs.CL},
  doi           = {10.48550/arXiv.2312.11805}
}

@inproceedings{hpe:vitpose:xu,
  author    = {Xu, Yufei and Zhang, Jing and Zhang, Qiming and Tao, Dacheng},
  title     = {{ViTPose}: Simple Vision Transformer Baselines for Human Pose Estimation},
  booktitle = {Advances in Neural Information Processing Systems},
  year      = {2022},
  volume    = {35},
  pages     = {38571--38584}
}

@inproceedings{ours:nlf,
  author    = {S{\'a}r{\'a}ndi, Istv{\'a}n and Pons-Moll, Gerard},
  title     = {Neural Localizer Fields for Continuous 3{D} Human Pose and Shape Estimation},
  booktitle = {Advances in Neural Information Processing Systems},
  year      = {2024},
  volume    = {37},
  pages     = {140032--140065},
  doi       = {10.52202/079017-4444}
}

@inproceedings{ours:nugget,
  author    = {K{\"a}s, Stephanie and Burenko, Anton and Markert, Louis and {\c{C}}ulha, Onur Alp and Mack, Dennis and Linder, Timm and Leibe, Bastian},
  title     = {How Do Foundation Models Compare to Skeleton-Based Approaches for Gesture Recognition in Human--Robot Interaction?},
  booktitle = {2025 34th IEEE International Conference on Robot and Human Interactive Communication (RO-MAN)},
  year      = {2025},
  pages     = {477--484},
  doi       = {10.1109/RO-MAN63969.2025.11217528}
}

@inproceedings{det:rf-detr:robinson,
  author    = {Robinson, Isaac and Robicheaux, Peter and Popov, Matvei and Ramanan, Deva and Peri, Neehar},
  title     = {{RF-DETR}: Neural Architecture Search for Real-Time Detection Transformers},
  booktitle = {The Fourteenth International Conference on Learning Representations},
  year      = {2026},
  %url       = {https://openreview.net/forum?id=qHm5GePxTh}
}

@article{hpe:openpose:cao,
  author  = {Cao, Zhe and Hidalgo, Gines and Simon, Tomas and Wei, Shih-En and Sheikh, Yaser},
  title   = {{OpenPose}: Realtime Multi-Person 2{D} Pose Estimation Using Part Affinity Fields},
  journal = {IEEE Transactions on Pattern Analysis and Machine Intelligence},
  year    = {2021},
  volume  = {43},
  number  = {1},
  pages   = {172--186},
  doi     = {10.1109/TPAMI.2019.2929257}
}

@inproceedings{hpe:hrnet:sun,
  author    = {Sun, Ke and Xiao, Bin and Liu, Dong and Wang, Jingdong},
  title     = {Deep High-Resolution Representation Learning for Human Pose Estimation},
  booktitle = {2019 IEEE/CVF Conference on Computer Vision and Pattern Recognition (CVPR)},
  year      = {2019},
  pages     = {5693--5703},
  doi       = {10.1109/CVPR.2019.00584}
}

@article{data:spacesuitpose:ouyang,
  author  = {Ouyang, Jingxuan and Xie, Wentao and Xu, Guangsheng and Li, Chujun and Gan, Shuwei and Zhang, Xiaohu and Yang, Xia and Li, You and Jiang, Changhua},
  title   = {{SpacesuitPose}: Deep Learning-Based Spacesuit Pose Estimation in Extravehicular Activities from Monocular Images},
  journal = {Acta Astronautica},
  year    = {2025},
  volume  = {234},
  pages   = {387--397},
  doi     = {10.1016/j.actaastro.2025.02.015}
}

@article{hpe:ropenpose:wu,
  author  = {Wu, Edmond Q. and Tang, Zhi-Ri and Xiong, Pengwen and Wei, Chuan-Feng and Song, Aiguo and Zhu, Li-Min},
  title   = {{ROpenPose}: A Rapider {OpenPose} Model for Astronaut Operation Attitude Detection},
  journal = {IEEE Transactions on Industrial Electronics},
  year    = {2022},
  volume  = {69},
  number  = {1},
  pages   = {1043--1052},
  doi     = {10.1109/TIE.2020.3048285}
}

@article{hpe:h3-astroposenet:xun,
  author  = {Xun, Ruiqi and Zhang, Zhang and Chang, Liang},
  title   = {{{H}$^3$-AstronautPoseNet}: A Multiview 3{D} Astronaut Pose Estimation System for Spacecraft},
  journal = {Neurocomputing},
  year    = {2025},
  volume  = {650},
  pages   = {130877},
  doi     = {10.1016/j.neucom.2025.130877}
}

@inproceedings{data:spacewalk-18:tang,
  author    = {Tang, Zitian and Krishnan, Rohan Myer and Yu, Zhiqiu and Sun, Chen},
  title     = {{Spacewalk-18}: A Benchmark for Multimodal and Long-Form Procedural Video Understanding in Novel Domains},
  booktitle = {2026 IEEE/CVF Winter Conference on Applications of Computer Vision (WACV)},
  year      = {2026},
  pages     = {4472--4482},
  %url       = {https://openaccess.thecvf.com/content/WACV2026/html/Tang_Spacewalk-18_A_Benchmark_for_Multimodal_and_Long-form_Procedural_Video_Understanding_WACV_2026_paper.html}
}

@InProceedings{data:coco:lin,
  author    = "Lin, Tsung-Yi and Maire, Michael and Belongie, Serge and
               Hays, James and Perona, Pietro and Ramanan, Deva and
               Doll{\'a}r, Piotr and Zitnick, C. Lawrence",
  editor    = "Fleet, David and Pajdla, Tomas and Schiele, Bernt and
               Tuytelaars, Tinne",
  title     = "Microsoft {COCO}: Common Objects in Context",
  booktitle = "Computer Vision -- {ECCV} 2014",
  series    = "Lecture Notes in Computer Science",
  volume    = 8693,
  publisher = "Springer International Publishing",
  address   = "Cham",
  pages     = "740--755",
  year      = 2014,
  doi       = "10.1007/978-3-319-10602-1_48",
  isbn      = "978-3-319-10602-1"
}

@inproceedings{data:egoexo4d:grauman,
  author    = {Grauman, Kristen and Westbury, Andrew and Torresani, Lorenzo and others},
  title     = {{Ego-Exo4D}: Understanding Skilled Human Activity from First- and Third-Person Perspectives},
  booktitle = {2024 IEEE/CVF Conference on Computer Vision and Pattern Recognition (CVPR)},
  year      = {2024},
  pages     = {19383--19400},
  doi       = {10.1109/CVPR52733.2024.01834}
}

@article{hpe:astropose:liu,
  author  = {Liu, Zibin and Li, You and Wang, Chunhui and Liu, Liang and Guan, Banglei and Shang, Yang and Yu, Qifeng},
  title   = {{AstroPose}: Astronaut Pose Estimation Using a Monocular Camera During Extravehicular Activities},
  journal = {Science China Technological Sciences},
  year    = {2024},
  volume  = {67},
  number  = {6},
  pages   = {1933--1945},
  doi     = {10.1007/s11431-023-2599-8}
}

@inproceedings{data:mupose:cardoso,
  author     = {{De Souza Cardoso}, Lu{\'i}s Fernando and Schwandt, Tobias and Broll, Wolfgang},
  title     = {{$\mu$Pose}: Synthetic Dataset for Human Pose Estimation in Microgravity Environments},
  booktitle = {2024 IEEE Conference on Artificial Intelligence (CAI)},
  year      = {2024},
  pages     = {1557--1562},
  doi       = {10.1109/CAI59869.2024.00282}
}

@misc{det:yolo26:jocher,
  author        = {Jocher, Glenn and Qiu, Jing and Liu, Mengyu and Lyu, Shuai and Akyon, Fatih Cagatay and Kalfaoglu, Muhammet Esat},
  title         = {Ultralytics {YOLO26}: Unified Real-Time End-to-End Vision Models},
  year          = {2026},
  eprint        = {2606.03748},
  archiveprefix = {arXiv},
  primaryclass  = {cs.CV},
  doi           = {10.48550/arXiv.2606.03748}
}

@inproceedings{det:yolo:redmon,
  author    = {Redmon, Joseph and Divvala, Santosh and Girshick, Ross and Farhadi, Ali},
  title     = {You Only Look Once: Unified, Real-Time Object Detection},
  booktitle = {2016 IEEE Conference on Computer Vision and Pattern Recognition (CVPR)},
  year      = {2016},
  pages     = {779--788},
  doi       = {10.1109/CVPR.2016.91}
}

@InProceedings{det:grounding-dino:liu,
  author    = "Liu, Shilong and Zeng, Zhaoyang and Ren, Tianhe and Li, Feng and
               Zhang, Hao and Yang, Jie and Jiang, Qing and Li, Chunyuan and
               Yang, Jianwei and Su, Hang and Zhu, Jun and Zhang, Lei",
  editor    = "Leonardis, Ale{\v{s}} and Ricci, Elisa and Roth, Stefan and
               Russakovsky, Olga and Sattler, Torsten and Varol, G{\"u}l",
  title     = "Grounding {DINO}: Marrying {DINO} with Grounded Pre-Training for Open-Set Object Detection",
  booktitle = "Computer Vision -- {ECCV} 2024",
  series    = "Lecture Notes in Computer Science",
  volume    = 15105,
  publisher = "Springer Nature Switzerland",
  address   = "Cham",
  pages     = "38--55",
  year      = 2025,
  doi       = "10.1007/978-3-031-72970-6_3",
  isbn      = "978-3-031-72970-6"
}

@techreport{Abercromby2006ReachEnvelope,
  author      = {Abercromby, Andrew F. J. and Thaxton, Sherry S. and Onady, Elizabeth A. and Rajulu, Sudhakar L.},
  title       = {Reach Envelope and Field of Vision Quantification in Mark {III} Space Suit Using Delaunay Triangulation},
  institution = {National Aeronautics and Space Administration},
  number      = {NASA/TP-2006-213729},
  year        = {2006}
}

@inproceedings{Kobrick2018SpacesuitROM,
  author    = {Kobrick, Ryan L. and Lopac, Nicholas and Schuman, Jenifer and Covello, Chase and French, John and Gould, Angelica and Meyer, Maximilian and Southern, Theodore and Lones, Jazmyne and Ehrlich, Joshua W.},
  title     = {Spacesuit Range of Motion Investigations Using Video and Motion Capture Systems at Spaceflight Analogue Expeditions and within the {ERAU S.U.I.T. Lab}},
  booktitle = {48th International Conference on Environmental Systems},
  year      = {2018},
  address   = {Albuquerque, NM, USA},
  note      = {Paper ICES-2018-189},
  %url       = {https://commons.erau.edu/publication/1436/}
}

@inproceedings{Kobrick2019FinalFrontierROM,
  author    = {Kobrick, Ryan L. and Lopac, Nicholas and Covello, Chase and Banner, Benjamin and Southern, Theodore and Moiseev, Nikolay},
  title     = {Range of Motion Evaluation of a Final Frontier Design {IVA} Spacesuit Using Motion Capture},
  booktitle = {49th International Conference on Environmental Systems},
  year      = {2019},
  address   = {Boston, MA, USA},
  note      = {Paper ICES-2019-99},
  %url       = {https://commons.erau.edu/publication/1435/}
}

@article{Doule2021IVASpacesuit,
  author  = {Doule, Ondrej and Kobrick, Ryan L. and Crisman, Keith and Skuhersky, Max and Lopac, Nicholas and Fornito, II, Michael J. and Covello, Chase and Banner, Benjamin C.},
  title   = {{IVA} Spacesuit for Commercial Spaceflight: Upper Body Motion Envelope Analysis},
  journal = {Acta Astronautica},
  year    = {2021},
  volume  = {186},
  pages   = {523--532},
  doi     = {https://doi.org/10.1016/j.actaastro.2021.05.026}
}

@inproceedings{Kobrick2012IMUSpacesuit,
  author    = {Kobrick, Ryan L. and Carr, Christopher E. and Meyen, Forrest and Domingues, Ana R. and Newman, Dava J. and Jacobs, Steven},
  title     = {Using Inertial Measurement Units for Measuring Spacesuit Mobility and Work Envelope Capability for Intravehicular and Extravehicular Activities},
  booktitle = {63rd International Astronautical Congress},
  year      = {2012},
  address   = {Naples, Italy},
  note      = {Paper IAC-12-A1.6.6}
}

@article{Fineman2018SpacesuitFit,
  author  = {Fineman, Richard A. and McGrath, Timothy M. and Kelty-Stephen, Damian G. and Abercromby, Andrew F. J. and Stirling, Leia A.},
  title   = {Objective Metrics Quantifying Fit and Performance in Spacesuit Assemblies},
  journal = {Aerospace Medicine and Human Performance},
  year    = {2018},
  volume  = {89},
  number  = {11},
  pages   = {985--995},
  doi     = {10.3357/AMHP.4937.2018}
}
\addcontentsline{toc}{section}{References}

\end{document}